\documentclass{article}
\usepackage{spconf}

\usepackage[english]{babel}
\usepackage{ifpdf}
\usepackage{cite} 
\usepackage{url}
\usepackage{hyperref}
\ifpdf
	\usepackage[pdftex]{graphicx}
	\graphicspath{{./figures/}}
\else
	\usepackage[dvips]{graphicx}
	\graphicspath{./figures/}
\fi
\usepackage{color}
\usepackage{pgf, tikz, pgfplots}
\usetikzlibrary{shapes, arrows, automata}
\usepackage{caption} 
\usepackage{subcaption} 
\usepackage{amsmath}
\usepackage{amsfonts, amssymb, amsthm}
\usepackage{mathrsfs}
\usepackage{upgreek}
\usepackage{algorithm,algpseudocode}
\usepackage{enumerate}
\usepackage{multirow}
\usepackage{rotating}
\usepackage{needspace}

\input{mysymbol.sty}

\newtheorem{definition}{\hspace{0pt}\bf Definition}

\newcommand{\QED}{\hfill\ensuremath{\blacksquare}}

\newcommand{\neighborhood}[2]{\mathcal{N}_{#1}^{#2}}
\newcommand{\extneighborhood}[2]{\overline{\mathcal{N}}_{#1}^{#2}}

\newcommand{\nmedian}[2]{{#1}\mbox{\textbf{-med}}_{\mathcal{G}}(\mathbf{{#2}})}

\newcommand{\wnmedian}[2]{{#1}\mbox{\textbf{-}}\mbox{\textbf{med}}_{\mathcal{G}}^{\mathbf{\omega}}(\mathbf{{#2}})}

\title{MEDIAN ACTIVATION FUNCTIONS FOR GRAPH NEURAL NETWORKS}
\name{Luana Ruiz$^{\dag}$, Fernando Gama$^{\dag}$, Antonio G. Marques$^{\ast}$ and Alejandro Ribeiro$^{\dag}$\thanks{Supported by USA NSF CCF 1717120 and ARO W911NF1710438, and Spanish MINECO TEC2016-75361-R. 
		}
}
\address{\dag \ Department of Electrical and Systems Engineering, University of Pennsylvania, Philadelphia, USA\\
$\ast$ \ Department of Signal Theory and Communications, King Juan Carlos University, Madrid, Spain
}
\begin{document}
%
\maketitle
\begin{abstract}
Graph neural networks (GNNs) have been shown to replicate convolutional neural networks' (CNNs) superior performance in many problems involving graphs. By replacing regular convolutions with linear shift-invariant graph filters (LSI-GFs), GNNs take into account the (irregular) structure of the graph and provide meaningful representations of network data. However, LSI-GFs fail to encode local \textit{nonlinear} graph signal behavior, and so do regular activation functions, which are nonlinear but pointwise. To address this issue, we propose median activation functions with support on graph neighborhoods instead of individual nodes. A GNN architecture with a trainable multirresolution version of this activation function is then tested on synthetic and real-word datasets, where we show that median activation functions can improve GNN capacity with marginal increase in complexity.
\end{abstract}
\begin{keywords}
graph neural networks, big data, activation functions, graph signal processing, median filters
\end{keywords}
\section{Introduction}
\label{sec:intro}

Modern problems of social and economic interest such as identifying the origin of epidemics on transportation networks and deciding which products to market to users of a social network have motivated the development of an extensive framework to model and analyze big data on networks \cite{Jackson08-Social}. In particular, many of these tasks require information processing architectures that are adapted to the inherently irregular structure of graphs. Graph convolutional neural networks (GNNs) are one such example. GNNs are the extension of convolutional neural networks (CNNs) to the graph domain and, like them, have demonstrated remarkable performance in multiple classification and regression tasks \cite{gama18-gnnarchit,defferrard17-cnngraphs,kipf17-classifgcnn, bruna14-deepspectralnetworks, ying18-pinterest}. They are good at generating graph data representations because their convolutional layers are based off of linear-shift invariant graph filters (LSI-GFs) that encode the irregular structure of the underlying graph into the filtered signals \cite{segarra17-linear}. 

Naturally, GNNs give good approximations of more than just linear functions, and the reason for that is the fact that each GNN layer composes the LSI-GFs with a nonlinear operator. In NNs, this operator is called the activation function, and common choices for it are the sigmoid, the $\tanh$ and the rectified linear unit (ReLu) \cite{goodfellow16-deeplearn}. Because these are all pointwise operations, in the GNN they can be computed at each node individually, without looking at the rest of the graph. While separating the local mixing from the non-linearities may render the analysis and implementation of the NN easier, architectures that allow for non-scalar nonlinear activation functions (i.e., activation functions whose input is a vector collecting the features in the neighborhood of a given node) are more general, either because they are better proxies of the actual underlying data model or because they achieve the same representation capabilities with less parameters.

Building upon this observation, we introduce \textit{median activation functions} as activation functions whose output at each node depends on the values of the input at the neighboring nodes. These functions will provide smooth and analytically tractable nonlinear transformations which, in some cases, can be related to median graph filters \cite{segarra16-globalsip,segarra17-camsap}. The main focus of this paper is on the case where the activation function is a weighted sum of medians taken at different graph resolutions, whose linear weights can be trained together with the other linear parameters of the GNN. While parametrized activation function training \cite{Chandra04-Sigmoid, He15-Rect} and activation functions that adapt from graph training data \cite{Scardapane18-KernelActiv} have been explored in the existing literature, they are all pointwise and, as such, unable to learn local graph topologies in the way that trainable median activations do.
This ability is demonstrated in experiments performed on both synthetic and real-world datasets, where GNN architectures with median and ReLu activations are compared. Our results show that median activation functions can further improve GNN performance by improving capacity while only marginally increasing the number of trainable parameters. 

\section{Graph Neural Networks}
\label{sec:gnn}

Consider a training set $\ccalT = \{(\bbx_{i}, \bby_{i})\}_{i=1}^{|\ccalT|}$ with $|\ccalT|$ samples where $\bbx \in \ccalX$ are the input data and $\bby \in \ccalY$ are known target representations that we intend to learn from. By learning we mean that we want to obtain suitable representations $\hby \in \ccalY$ of unseen inputs $\bbx' \notin \ccalT$; to accomplish this, we train a model $f: \ccalX \to \ccalY$ such that $\hby = f(\bbx)$ and $f$ minimizes the value of a loss function $\ccalL(\hby,\bby_{i})$ on $\ccalT$. We make the assumption that the unseen inputs $\bbx' \notin \ccalT$ are drawn from the same or from a similar distribution as the $\bbx \in \ccalT$, in which case it is reasonable to expect $f$ to output accurate representations of $\bbx'$.

NNs are the architecture of our choice in coming up with $f$. They are built from the concatenation of a series of computational layers which, in turn, perform two basic operations: a linear transform and a pointwise nonlinearity that is also referred to as activation function. Layer $\ell$ will output $\bbx_\ell$ given by
\begin{equation} \label{eqn:nn}
\bbx_{\ell} = \sigma_{\ell}(T_{\bbA_\ell}(\bbx_{\ell-1})) = \sigma_{\ell}(\bbA_{\ell} \bbx_{\ell-1}),\ \ell = 1,\ldots,L
\end{equation}
where $\bbx_\ell \in \ccalX_\ell$, $T_{\bbA_\ell} : \ccalX_{\ell-1} \to \ccalX_\ell$ is the linear transform and $\sigma_\ell : \ccalX_\ell \to \ccalX_\ell$ is the pointwise nonlinearity. We take the input $\bbx_0$ and the output $\bbx_L$ to be $\bbx$ and $\hby$ respectively. 

In this paper, we are interested in the case where $\bbx$ is a graph signal. A graph problem of this sort would be, for instance, the task of predicting a power outage in a given city based on information collected at the region's meteorological stations, modeled as the vertices of a geodesic graph \cite{Owerko18-PowerOutages}. Graph signals are represented as vectors $\bbx \in \mbR^N$; the individual components $[\bbx]_i$ are the values of the signal at each node $i$ of the graph, which we denote as $\ccalG$. For our purposes, $\ccalG$ is fully determined by the triplet $\ccalG=(\ccalV,\ccalE,\ccalW$), where $\ccalV = \{1,\ldots,N\}$ is the set of nodes,  $\ccalE \subseteq \ccalV \times \ccalV$ is the set of edges and $\ccalW : \ccalE \to \mbR$ is the weight function of the graph, $\ccalW(i,j) = w_{ij}$. Additionally, $\ccalG$ can be represented in matrix form: the graph adjacency matrix $\bbW \in \reals^{N \times N}$, where $[\bbW]_{ij} := w_{ji}$, and the unnormalized graph Laplacian $\bbL := \bbD - \bbW$, with $\bbD = \diag(\bbW\mathbf{1})$ the degree matrix of $\ccalG$, are the two most common representations. 

Both $\bbW$ and $\bbL$ are examples of the so-called \textit{graph shift operator} (GSO) of the graph. A GSO $\bbS \in \mbR^{N \times N}$ of $\ccalG$ is defined as any matrix satisfying $[\bbS]_{ij} \neq 0$ if and only if $(i,j) \in \ccalE$ or $i = j$. In this paper, the GSO $\bbM \in \{0,1\}^{N \times N}$ such that $[\bbM]_{ij} = 1$ for every $(i,j) \in \ccalE$ is a particularly useful one. Its powers $\bbM^r$, $r \geq 0$, allow us to define the $r$-hop neighborhood of a node $i$ as $\neighborhood{i}{r}$, 
%
that is, the set of nodes that are exactly at $r$ hops from $i$.

Since we are interested in graph signals, the generic architecture described by \eqref{eqn:nn} has to be adapted to process this type of data. In GNNs, this is done by imposing that $T_\bbA(\bbx)$ be an LSI-GF, where $\bbA$ in \eqref{eqn:nn} takes the form of $\bbH : \mbR^N \to \mbR^N$ such that
\begin{equation} \label{eqn:lsigf}
\bbH:= \sum_{k=0}^{K-1}h_k\bbS^k,
\end{equation}
with $h_0,\ldots,h_{K-1}$ the filter coefficients and $K$ the number of filter taps. 

Another differentiating characteristic of GNNs is the number of filters that they have per layer. Like CNNs designed for image processing applications, GNNs also rely on different filters to extract distinct local features at each layer. Each of these filters outputs a linear feature that could either be combined with others through a summarizing operation or be considered individually, at the price of increasing the subsequent layer's dimensionality. 

Extending the activation functions $\sigma$ to GNNs is immediate because these operations are inherently pointwise. Applying $\sigma$ to an arbitrary graph signal $\bby$ yields $[\sigma(\bby)]_i = \sigma([\bby]_i)$; this means that the outcome of the activation function at node $i$ depends solely on the value of the input signal at this node. In this paper, we argue that, since the overall goal is to merge information across nodes in a nonlinear fashion, considering only scalar activation functions can be too restrictive.
This idea --- that it takes nonlinear operators to capture nonlinear relationships in data --- is not new and has for instance motivated the authors of \cite{shen18-online} to come up with a dynamic-nonlinear method for online identification of graph topologies that does better than traditional linear structural vector autoregressive models.

In this context, we put forward \textit{median activation functions} as local operators that provide a nonlinear yet relatively smooth measure of central tendency in variously sized regions of a graph around each node. The main reasons for choosing this family of nonlinearities are the fact that they preserve tractability (both computationally and analytically) and are effective in many inference and signal processing tasks. With median activations, we expect to provide an alternative, more efficient way of increasing GNNs' overall capacity \cite{goodfellow16-deeplearn} to wider or deeper layered architectures, that risk overfitting when parameters are added in excess.

\section{Median activation functions}
\label{sec:median}

The median activation functions introduced in this paper differ from conventional activation functions in two ways. The first and most important difference is in their support, which is now a graph neighborhood instead of a single node. The second is the choice of nonlinear operation to perform --- the median graph filter. Indeed, the activation function discussed in the next paragraph is simply a $r$-hop median graph filter where, for every node $i$, the filter window is defined as the node's extended $r$-hop neighborhood $\extneighborhood{i}{r}$,
\begin{equation} \label {eqn:R_neigh}
 \extneighborhood{i}{r}:= \bigcup\limits_{k=0}^{r} \neighborhood{i}{k}.
\end{equation}
To make the distinction between the two activation functions introduced in this paper, we will refer to this first one as the \textit{static median activation}. We introduce it as a first step to later define a more interesting operator, namely the \textit{dynamic median activation}. The static median activation function is defined as follows.

%
\begin{definition} \label {defn:graph_med} 
Consider a graph $\ccalG = (\ccalV, \ccalE, \ccalW)$, a graph signal $\bbx \in \mbR^N$ on $\ccalG$ and the number of hops $r \in \mbN$. We define the $r$-hop median graph filter as $\nmedian{r}{x}: \mbR^N \to \mbR^N$ such that
\begin{equation} \label {eqn:graph_med}
\begin{split}
&[\nmedian{r}{x}]_i := \mbox{med}\{x_{j_1}, x_{j_2},...,x_{j_{N^r_i}}\} \mbox{ with} \\
&N^r_i = |\extneighborhood{i}{r}| \mbox{ and } j_1, j_2,...,j_{N^r_i} \in \extneighborhood{i}{r}.
\end{split}
\end{equation}
For $r>0$ and $\ccalG$ without unitary components (isolated nodes), $\nmedian{r}{x}$ defines the \textbf{static median activation function} for GNN architectures.
\end{definition}

What this definition conveys is that the static median activation function takes a graph signal as input, fetches each node's extended $r$-hop neighborhood, and computes $N$ medians of the signal in these neighborhoods to output another signal on the same graph. If $N^r_i$ in equation \eqref{eqn:graph_med} is even, we stipulate $[\nmedian{r}{x}]_i = x_{[N^r_i/2+1]}$, that is, the median is given by the $(N^r_i/2+1)$th order statistics of $x_{j_1},\ldots,x_{j_{N^r_i}}$. We point out that, as long as $r>0$, the operations in \eqref{eqn:graph_med} yield nonlinear activation functions.

While the graph filter from Definition \ref{defn:graph_med} has the advantage of enlarging the activation function support beyond a single point or node, the $r$-hop static median activation is only able to capture nonlinear relationships in the $r$-hop neighborhood of each node. However, the behavior of the graph signal at a given node might be more complex than what this function is able to seize; for instance, it could be as related to the behavior of the signal on closer nodes (within the $r$-hop neighborhood) as to the signal dynamics on more distant but highly influential nodes (e.g. nodes with high centrality). The dynamic median activation function presented in Definition \ref{defn:w_graph_med} addresses this issue by taking into account the medians in node neighborhoods of increasing size (up to $R$ hops, where $R$ is the median filter's reach) and weighing them linearly. We define it as follows.

%
\begin{definition} \label {defn:w_graph_med} 
Consider a graph $\ccalG = (\ccalV, \ccalE, \ccalW)$, a graph signal $\bbx \in \mbR^N$ defined on $\ccalG$, the filter reach $R \in \mbN$ and the linear weight vector $\omega \in \mbR^{R+1}$. We define the $R$-hop weighted graph filter $\wnmedian{R}{x}: \mbR^N \to \mbR^N$ as
\begin{equation} \label {eqn:w_graph_med}
\begin{split}
 &[\wnmedian{R}{x}]_i := \omega^{\top} \bbz_i, \\
 &\bbz_i \in \mbR^{R+1} \mbox{ and } [\bbz_i]_r = [\nmedian{(r)}{x}]_i
\end{split}
\end{equation}
with $r=0,\ldots,R$ and $i=1,\ldots,N$. 
For $R>0$ and $\ccalG$ without unitary components (isolated nodes), $\wnmedian{R}{x}$ defines the \textbf{dynamic median activation function} for GNN architectures.
\end{definition}

The introduction of linear weight variables in the above definition is desirable for two reasons. The first one is that they allow assigning different levels of importance to the median in each neighborhood. The second is the fact that, because these variables are linear, their gradients are easy to compute and so they can be updated via backpropagation. In other words, this activation function can be trained, which is why we call it \textit{dynamic}. It is worth pointing out that even though the dynamic median activation function is linear in the weights $\omega$, it is still nonlinear in the input graph signal $\bbx$ as long as $R>0$.

\section{Numerical Experiments}
\label{sec:sims}

\begin{figure}[t]

\begin{minipage}[t]{1.0\linewidth}
  \centering
  \centerline{\includegraphics[width=7cm]{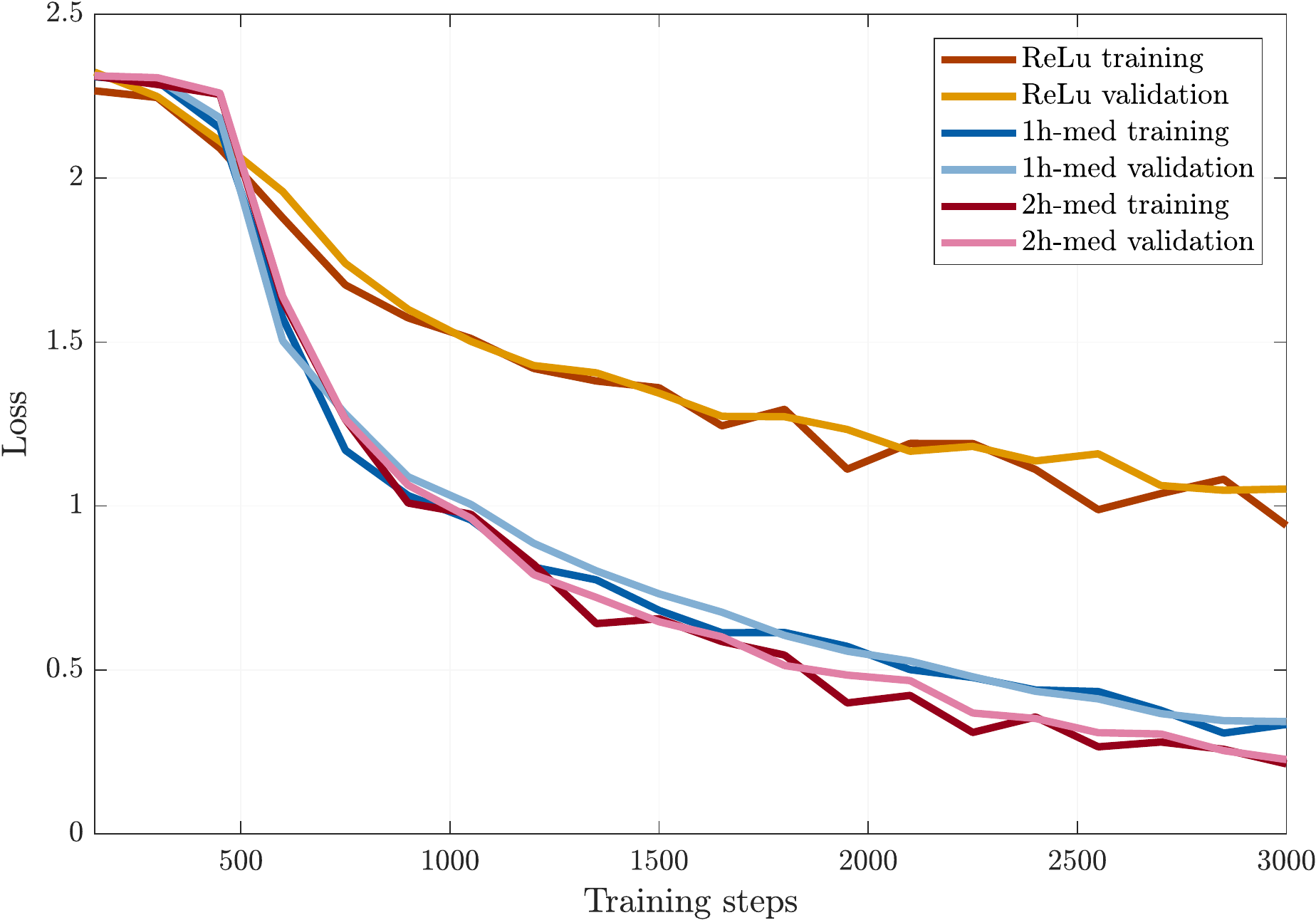}}
\end{minipage}
\caption{Training and validation losses in one round of simulation of the source localization problem on the Twitter graph, for the ReLu, 1-hop dynamic median and 2-hop dynamic median architectures .}
\label{fig:loss}
\end{figure}

In this section, we present simulation results obtained with GNN architectures using \textit{dynamic median activation functions} in two applications. In the first, we try to predict the source of a synthetic diffusion process on social networks; in the second, we attribute authorship of text excerpts to 19th century writers. Our architecture's performance is compared with that of a GNN that shares the same design parameters in the convolutional layer, except that the activation function is a (pointwise) ReLu.

The simulated GNN architectures all consist of a single convolutional layer followed by a fully connected softmax layer, which we use to compute the cross-entropy loss. The convolutional layer contains 32 filters with 5 filter taps each; each filter uses the adjacency matrix normalized by its largest eigenvalue as the GSO and outputs a single feature. The GNNs were trained without dropout using the ADAM algorithm, a variation of stochastic gradient descent that keeps an exponentially decaying average of past gradients, with decaying factors $\beta_{1} = 0.9$ and $\beta_{2} = 0.999$ \cite{kingma17-adam}. 

\subsection{Source Localization on Facebook and Twitter} 

In this experiment, we consider a graph $\ccalG$ with $N$ nodes and generate graph diffusion processes $\bbx(t)$ \cite{segarra16-deconv} with origin at an arbitrary node $c$. Suppose that $\ccalG$ has adjacency matrix $\bbW$ and that $\bbx(0) = \bbx_0$, where $\bbx_0$ is such that $[\bbx_0]_i = 1$ if and only if $i=c$ and 0 everywhere else. The diffused signals $\bbx(t)$, $t = 1,2,\ldots$, are generated as $\bbx(t)=\bbW^t\bbx_0$. By picking random values of $t$ and placing the corresponding $\bbx(t)$ in tuples $(\bbx(t),c)$, we can train a GNN that when presented with a new, unobserved process $\bbx'(t)$, is able to accurately predict its source $c'$. 

This experiment is performed in two graphs: a 234-node undirected graph formed by a single connected component of the Facebook friend network; and a directed ego network \cite{leskovec2016snap} from Twitter with 237 nodes and a single weakly connected component. These networks were built under the Stanford Network Analysis Project and can be found in \cite{snapnets}. Although the diffusion processes used to train the GNNs in this paper are synthetic, they are a good representation of how rumors spread on social networks, for instance \cite{ortega2018gsp}.

In both the Facebook and Twitter experiments, 3 GNNs were trained on the same randomly generated diffusion datasets at a time: one with ReLu activations, one with 1-hop dynamic median activations and one with 2-hop dynamic median activations. In total, ten rounds of simulations corresponding to ten different datasets were conducted for each problem.  We used 10,000 training and 200 test samples in every round, and training was done in batches of 100 samples with learning rate 0.001. In the Facebook problem, we restricted the number of classes to 5, corresponding to the 5 nodes with highest degree. For Twitter, we allowed for the 10 nodes with highest degree to originate the diffusion process. The Facebook GNNs were trained in 40 epochs, while the Twitter ones only needed 30 training epochs in total. 

Figure \ref{fig:loss} shows the training and validation losses for every analyzed architecture in one of the 10 rounds of simulations on the Twitter graph. In this realization, it is clear that architectures with median activation functions incur a dramatic increase in capacity and sensibly reduce the generalization error while only adding up to 3 extra parameters to the GNN (cf. \ref{tb:sourceloc}). In particular, the best results are obtained with the median activation function with largest reach ($R=2$). Average test accuracy results for both the Facebook and Twitter problems are reported in Table \ref{tb:sourceloc}. While in the Facebook graph the 1-hop and 2-hop median achieve a similar accuracy (with the 1-hop median actually performing marginally better due to the smaller standard deviation), in the Twitter graph the benefits of increasing the median reach $R$ are significant. This could be explained not only by the fact that one graph is undirected and the other is directed, but also because while the Facebook graph has average degree 34.25, the Twitter graph has average in and out-degrees of 11.45. As such, there is much more redundancy between 1-hop and 2-hop extended neighborhoods in the Facebook graph than in the Twitter graph.

\begin{table}[t]
\centering
\begin{tabular}{l|cc|c} \hline
		& \multicolumn{2}{|c|}{Average accuracy (\%)} & \\
Archit.    & Facebook  & Twitter & Parameters\\ \hline
ReLu		& $70.30 \pm 9.08$ & $62.10 \pm 11.91$ & $160$	\\
1h-med		& $\mathbf{79.25} \pm \mathbf{1.72}$ & $97.60 \pm 4.44$ & $162$  \\
2h-med	& $79.80 \pm 2.39$ & $\mathbf{99.2} \pm \mathbf{2.24}$ & $163$   \\ \hline
\end{tabular}
\caption{Test accuracy for ReLu, 1-hop median and 2-hop median architectures in the source localization problem. Average and standard deviation for 200 test samples over 10 rounds of simulations on both the Facebook and Twitter graphs. The number of parameters in the convolutional layer of each architecture is reported in the rightmost column.}
\label{tb:sourceloc}
\end{table} 

\subsection{Authorship Attribution of 19th Century Texts}

The second application in which we assess dynamic median activation function performance is a binary classification problem based on real data. In this experiment, the graphs are author word adjacency networks (WANs) where each function word (e.g. prepositions, articles, conjunctions etc.) is a node and function words are connected through edges weighted by frequency of co-appearance in texts that are known to have been written by the author in question \cite{eisen15-auth}. Because order of appearance matters, these graphs are directed. We consider WANs by two authors: Jacob Abbott and Robert Stevenson. On these graphs, graph signals are defined as the vectored frequencies of each function word in 1,000-word text excerpts. The text excerpts we consider come from a pool of twenty-one 19th century authors, including Abbott and Stevenson; then, for the GNN trained on Abbot's WAN, for example, the objective is to classify each excerpt as having being written by Abbott or not.

Here again we trained 3 GNNs for Abbott and 3 for Stevenson, one per activation function (ReLu, 1h-med, 2h-med). The number of parameters in the convolutional layer of each GNN were, respectively, 160, 162 and 163. For each author, we conducted 10 rounds of simulations varying both the WAN and the training and test set 80-20 splits. We consider 180 function words (nodes) for Abbott and 193 for Stevenson, and the resulting WANs are connected graphs.  Naturally, only the author's texts in the training set were used to build the WAN at each simulation round. For Abbott, we used 1,080 training and 268 test samples, and, for Stevenson, 1,436 and 356 respectively. All of these datasets are evenly balanced, meaning that half of the samples correspond to texts by the author in question (labeled +1, positive) and the other half are randomly picked from the 21-author pool (labeled -1, negative). Training was done in 10 epochs in batches of 20.

\begin{figure}[t]
    
    \begin{minipage}[t]{1.0\linewidth}
        \centering
        \centerline{\includegraphics[width=7cm]{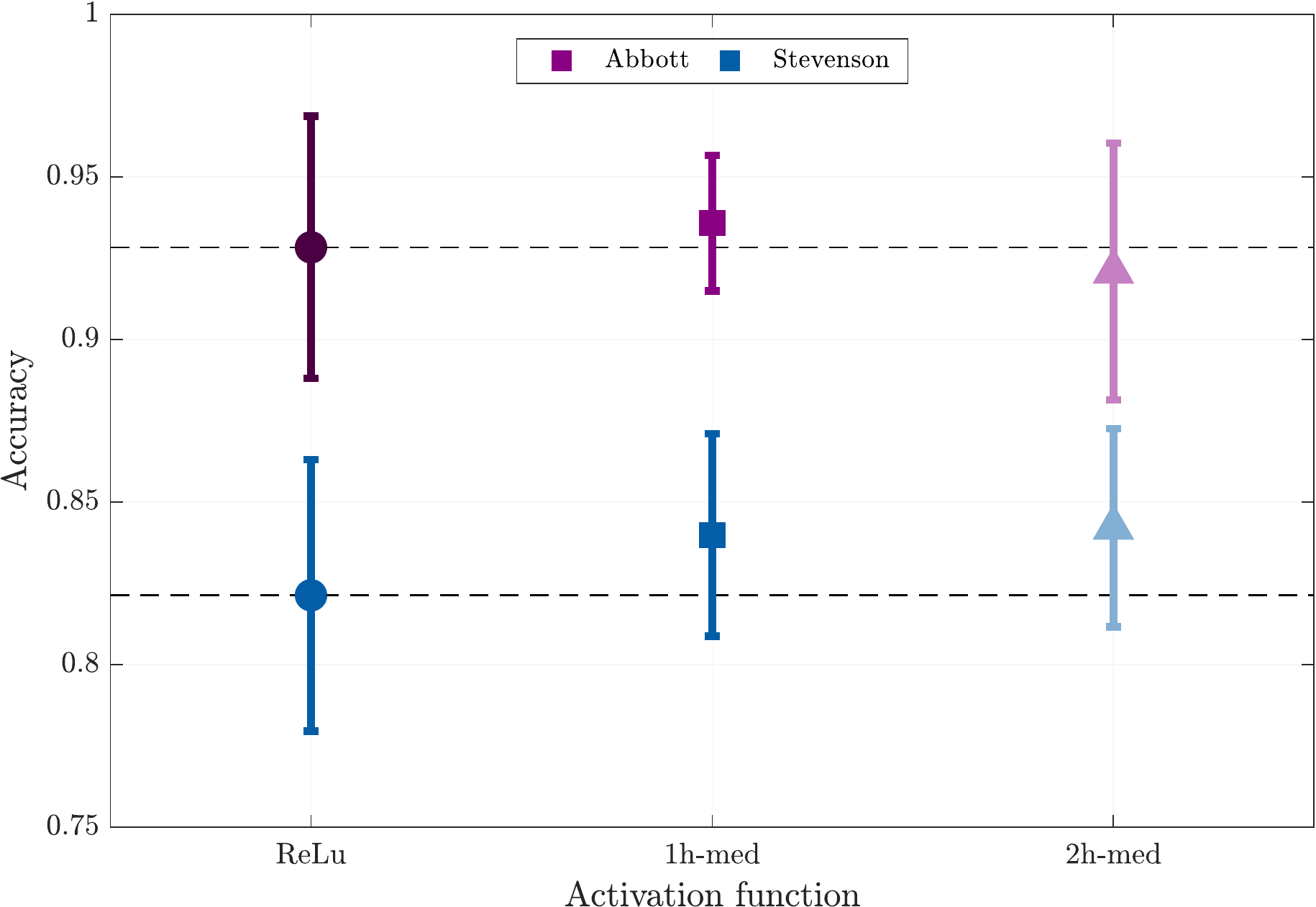}}
    \end{minipage}
    \caption{Test accuracy for ReLu, 1-hop median and 2-hop median architectures in the authorship attribution problem for Abbott and Stevenson. Average accuracy and standard deviation over 10 rounds of simulations.}
    \label{fig:auth}
\end{figure}

Average classification accuracies for both Abbott and Stevenson are shown in Figure \ref{fig:auth}. For Stevenson, the two median activation functions incur a significant improvement in average accuracy and consistency with respect to the ReLu, with around 7 additional texts being correctly classified on average. Another thing to note for this author is that performance increases with the activation function reach. For Abbott, the average accuracy improvement is more timid and comes from the 1-hop median, while the 2-hop median degrades performance w.r.t. the ReLu. Here, this is probably less related to growing neighborhood redundancies than it is to stylometric differences such as a particular author's average sentence length, for instance.

\section{Conclusions}
\label{sec:conclusions}

This work introduced static and dynamic median activation functions as an alternative to regular pointwise activation functions in GNNs. Their major advantage is extending activation function support to graph neighborhoods that go beyond individual nodes. This allows encoding any local nonlinear relationships existent in the graph and/or the graph signal, leading to increased GNN capacity. In particular, the dynamic median activation function takes the graph signal median in graph neighborhoods of increasing size and weighs them linearly. These weights can then be trained via backpropagation and thus yield a class of trainable activation functions. Numerical results in synthetic datasets showed that dynamic median activation functions are a better alternative to increasing GNN capacity than deepening or widening GNNs; they achieved low generalization error while keeping almost the same number of parameters as the GNN with all ReLu activations, thus decreasing the risk of overfitting. Additionally, we observed sensible classification accuracy improvements in both synthetic and real-world datasets.

\bibliographystyle{IEEEbib}
\bibliography{myIEEEabrv,bib-nonlinear}

\end{document}